\pdfoutput=1
\documentclass[11pt]{article}
\usepackage{ACL2023}
\usepackage{times}
\usepackage{latexsym}
\usepackage[T1]{fontenc}
\usepackage[utf8]{inputenc}
\usepackage{microtype}
\usepackage{inconsolata}

\usepackage{times,latexsym}
\usepackage{url}
\usepackage[T1]{fontenc}
\usepackage{breqn}

\usepackage{times}
\usepackage{color} 
\usepackage{latexsym}
\usepackage{subcaption}

\usepackage{amsmath,mathtools}
\usepackage{color,graphicx}
\usepackage{pgfplots}
\usepackage{adjustbox}
\usepackage{enumitem}

\usepackage{dblfloatfix}

\usepackage{titlesec}

\usepackage{xspace,mfirstuc,tabulary}

\usepackage{makecell}
\usepackage{multirow}
\usepackage{booktabs}
\usepackage{amsmath}
\usepackage{amsfonts}
\usepackage{mathtools}

\usepackage{graphicx}
\graphicspath{{figs/}}
\usepackage{xcolor}
\usepackage{placeins}

\newcommand{\quash}[1]{}

\title{Locally Measuring Cross-lingual Lexical Alignment: A Domain and Word Level Perspective}

\author{Taelin Karidi, Eitan Grossman, Omri Abend\\ 
Hebrew University of Jerusalem \\\texttt{\small  \{taelin.karidi,eitan.grossman,omri.abend\}@mail.huji.ac.il}}

\date{}

\begin{document}
\maketitle

\begin{abstract}

NLP research on aligning lexical representation spaces to one another has so far focused on aligning language spaces in their entirety. 
However, cognitive science has long focused on a local perspective, investigating whether translation equivalents truly share the same meaning or the extent that cultural and regional influences result in meaning variations. 
With recent technological advances and the increasing amounts of available data, the longstanding question of cross-lingual lexical alignment can now be approached in a more data-driven manner. However,  developing metrics for the task requires some methodology for comparing metric efficacy. We address this gap and present a methodology for analyzing both synthetic validations and a novel naturalistic validation using lexical gaps in the kinship domain.
We further propose new metrics, hitherto unexplored on this task, based on contextualized embeddings.\footnote{Our code and data is available at \url{https://github.com/tai314159/LEXI}.} 
Our analysis spans 16 diverse languages, demonstrating that there is substantial room for improvement with the use of newer language models. Our research paves the way for more accurate and nuanced cross-lingual lexical alignment methodologies and evaluation.

\end{abstract}

\section{Introduction}
\label{sec:intro}

Cross-lingual lexical semantic similarity can be approached from two complementary perspectives: \textit{local} and \textit{global}. 
The local perspective compares words or sets of words and characterizes how similarly meanings are lexicalized across languages \citep{wierzbicka1972semantic,majid2008,berlin1991basic,srinivasan2015concepts,Thompson_Lupyan20,Georgakopoulos21,Purves23}. For example, whether translation equivalents, such as the English \textit{green} and French \textit{vert}, encode the same meaning. In contrast, the global perspective focuses on how similar languages are as a whole, examining broader patterns, relationships, and structures within languages, rather than focusing on individual words. For instance, English and Bulgarian are both SVO languages, although the latter has a more flexible word-order structure. Such differences can influence their global alignment and the ability to transfer knowledge between them \citep{nikolaev2020fine,arviv2023improving}. 

\begin{figure}
    \centering
    \includegraphics[width=\columnwidth]{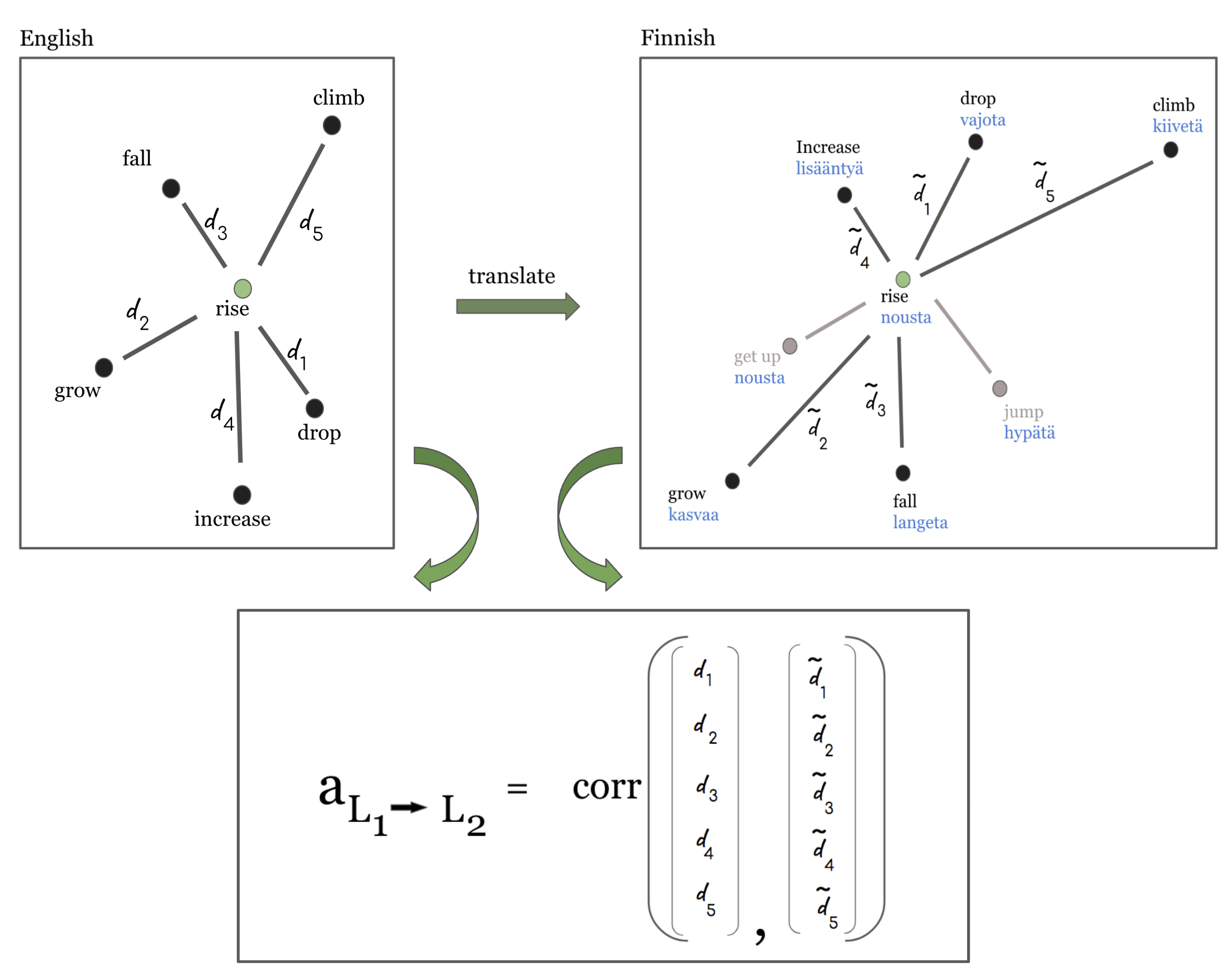}
    \caption{Distribution-based alignment. An illustration of the distribution-based alignment process for the word {\it rise} between English ($L_1$, upper left) and Finnish ($L_2$, upper right). First, we find the $k$ nearest neighbors of the word {\it rise} in $L_1$ (for ease of visualization $k=5$). We then translate the neighbors to $L_2$. The neighbors do not necessarily coincide with the nearest neighbors of the Finnish translation of {\it rise}. Here two of the nearest neighbors are not translations (marked in grey) and are ignored in the calculation). We then measure the correlation between the distance vectors. We repeat this in the opposite direction, $L_2 \to L_1$, and compute the average. }
    \label{fig:main_fig_dista_alignment}
\end{figure}


In the field of Natural Language Processing (NLP), most research on cross-lingual similarity focuses on global similarity measures \citep{Conneau17-muse,artetxe-etal-2018-robust,ruder2019survey,vulic-etal-2021-lexfit}, as they facilitate cross-lingual transfer in a multitude of tasks, such as machine translation and bilingual lexicon induction \citep{schuster2019cross,artetxe2019cross,eronen2023zero}. These approaches typically involve aligning entire vector spaces of different languages through methods such as linear transformations, aiming to create a unified semantic space where languages can be directly compared. 


The local perspective, although often overlooked in NLP, has long been a topic of interest in cognitive sciences and linguistics \citep{whorf56,fodor75,Wierzbicka_primes_universals,burns94,Snedeker2004,majid2008,croft2010,youn2016}.
To the extent that lexicons reflect the structure of human cognition, understanding how meaning is expressed across languages offers insight as to how humans categorize and represent the world.
Examples of such inquiries span various semantic {\it domains} (a way of grouping words together based on common aspects of meaning or function), including colors \citep{berlin1991basic} and emotions \citep{jackson18}.

Traditionally, in linguistic and cognitive research, comparing the meaning of words across languages involves methodologies and approaches that are less data-driven in nature, but rather prioritize in-depth, relatively small-scale exploration of meaning, such as descriptive comparisons \citep{wierzbicka1972semantic}, elicitation studies \citep{Barnett77,Tokowicz2002,Moldovan12,Allen2013,Purves23} and semantic maps \citep{haspelmath2003geometry,Croft22}. 

However, defining and operationalizing the notion of cross-lingual lexical similarity is notoriously difficult and controversial. The difficulty in defining lexical similarity has motivated a turn from theory-driven to data-driven approaches. Indeed, considerable recent work pursued data-driven approaches to the quantification of equivalency between word pairs in different languages \citep{Majid15,youn2016,jackson18,Georgakopoulos21}.  
While some studies employed NLP methods  \citep{thompson18,Thompson_Lupyan20,Rabinovich20,Beinborn20}, their application has been limited to static word representations and their validation have focused on converging evidence, i.e, getting the same (or similar) result in different means and looking for confounds. E.g, comparing to other tests from the cognitive science literature, such as picture naming \cite{multipic2018} or translation norms \citep{Tokowicz2002,Allen2013}. However, these test are only applicable to a very small set of languages and word lists. Moreover, in NLP, it is common to compare to some external reference point that we perceive is one of high quality. For brevity, we simply refer to it as ground truth (Section \S\ref{subsec:validation_lex_gaps}).

In this work, we focus on the challenging task of word-level semantic similarity across languages, addressing the question of how word meanings vary across languages.
We use a range of metrics designed to assess and compare the nuanced meanings of translation equivalents in different languages, and extend them in a novel way to include contextualized word representations (Section \ref{subsec:alignment_metrics}).

One of the main challenges in developing metrics to quantify differences in lexical semantics across languages is the difficulty to validate them, given that there are no available resources tailored to define ground truth in this area. 
To address this, we not only conduct extensive synthetic analyses of the metrics (Section \S\ref{subsec:validation_synthetic}) but also establish a novel validation method, adapting a newly compiled linguistic resource of lexical gaps in the kinship domain \citep[Section \S\ref{subsec:validation_lex_gaps};][]{Khishigsuren23}.

We perform a detailed comparison between the various metrics at two levels of granularity: word-level and domain level. 
We also analyze what features affect the alignment, using a combination of lexical and environmental features.
We show that at the domain-level there is  substantial similarity between the methods and that it offers a more stable level for such analysis. Moreover, while the contextualised embeddings (CE's) based metrics are substantially correlated with our naturalistic validation, the other methods are not, suggesting that there is definitely room for improvement in this area, using newer models.

To summarize our contributions we (1) formulate the question of cross-lingual lexical similarity as an NLP task; (2) compare and analyze various metrics for this task; (3) introduce novel metrics based on contextualised word embedding; (4) offer a comprehensive validation suite to support our findings, including a novel validation method against a high-quality linguistic resource specifically tailored to the kinship domain. 


\section{Related Work}
\label{sec:related_work}

A multitude of different approaches for computing distributional similarity have been explored in NLP, of which we select a number of representative examples.  
Distributional metrics can be classified based on whether they employ a joint space for the embeddings for the languages in question, or whether the spaces are trained monolingually and then aligned \cite{artetxe-etal-2018-robust,Conneau17-muse}.\footnote{We are aware of one study of cross-lingual lexical comparison that used global alignment to project languages to a shared space, and defined the degree of alignment between a translation pair to be the distance of the image of one word to the embedding of the other \citep{Rabinovich20}. 
However, due to the reasons stated above, we do not consider their approach in this paper.} The latter approaches have been  a key facilitator of cross-lingual transfer in NLP and are 
especially important in low-resource settings.  
However, for identifying patterns of divergence and convergence in the usage of specific words and domains, this approach is suboptimal.\footnote{To measure cross-lingual lexical alignment using global alignment, it is natural to define the distance (i.e., alignment) between a translation pair as their cosine distance in the shared (aligned) space. This definition is employed to (1) align the source and target language spaces, and (2) evaluate the accuracy of the alignment.} Globally optimal alignment (one that minimizes the distance between the image of one language in the space of another language) may distort the alignment of some words subsets, in the interest of improving the alignment of other, larger word sets.\footnote{Preliminary experiments conducted across various languages have shown that these methods do not correlate with other measures or with the validations we propose.} 

Local alignment, or the extent to which translation pairs like English \textit{home} and Spanish \textit{casa}, hold a similar meaning across languages, is a well studied open question in cognitive science \citep{berlin1991basic,majid2008,Majid15,youn2016,jackson18}, that had only recently been approached with NLP tools \citep{thompson18,Thompson_Lupyan20}. However, existing metrics are limited to static word embeddings and do not accommodate newer models that support contextualization.

Additionally, understanding the variability in meaning across languages can provide valuable insight into cultural differences, revealing how various societies conceptualize their unique experiences and worldviews \citep{qi2017reconstructing,khalilia2023lexical,shioiri2023cross,tjuka2024universal}. 
This line of research aligns with the recent surge in studies concerning multicultural knowledge in LLMs, which assess whether models like GPT variants or multimodal LMs possess diverse cultural knowledge or exhibit biases favoring Western cultures \citep{hershcovich2022challenges,havaldar2023multilingual,ventura2023navigating,cao2023assessing}.


\begin{table}[!t]
\centering
\footnotesize
\renewcommand{\tabcolsep}{0.15cm}
\begin{tabular}{cccc}
\toprule

&
{\sc English Form}& 
{\sc Concept} &
{\sc Domain} \\

\midrule

\multirow{3}{*}{\rotatebox[origin=c]{90}} & mother & mutter::N & Kinship \\
& mind & verstand::N & Cognition \\
& go & gehen::V & Motion \\
& today & heute::ADV & Time \\
& towel & Handtuch::N & Clothing \\
& business & Geschäft::N & Modern world\\
& hold & halten::V &Possession \\
& one & eins::NUM & Quantity \\
& floor & Fußboden::N & The house \\
& flower & Blume::N & Agriculture \\
& middle & Mitte::N & Spatial relations \\
& happiness & Glück::N & Emotions \\
& horse & Pferd::N & Animals \\
& red & rot::A & Sense perception \\ 
& break & brechen::V & Basic actions \\
& church & Kirche::N & Social \\
& write & schreiben::V & Language \\ 
& bread & Brot::N & Food and drink \\
& skin & Haut::N & The body \\

\bottomrule
\end{tabular}
\caption{Concepts and their domains. Examples of concepts, labled according to the NEL dataset (\S\ref{sec:exp_setup}). ``Domain'' designates the semantic domain the concept belongs to, and ``English Form'' designates the lexicalization of each concept in English. For space considerations, ``Clothing'' denotes ``Clothing and grooming'',  ``Agriculture'' denoted ``Agriculture and Vegitation'', ``Basic Actions'' denotes ``Basic actions and technology'', ``Social'' denotes ``Social and political relations'', ``Emotions'' denotes ``Emotions and values'' and ``Language'' denotes ``Speech and language''.}
\label{tab:concepts_domains}
\end{table}

\section{Experimental Setup} \label{sec:exp_setup}

We provide a brief description of our experimental setup, with full details available in Appendix \S\ref{app:sec:exp_setup}.

\paragraph{Languages.} We perform our analysis on a diverse set of $16$ languages, spanning 7 different top-level language families from many geographical areas across Eurasia: English (eng), French (fra), Italian (ita), German (deu), Dutch (nld), Spanish (spa), Polish (pol), Finnish (fin), Estonian (est), Turkish (tur), Chinese (chn), Korean (kor), Japanese (jap), Hebrew (heb), Hindi (hin) and Arabic (arb). 

\paragraph{Data.} We conduct our analysis on the NorthEuraLex (NEL) dataset, a lexical resource compiled from dictionaries and other linguistic resources, such as concept lists, available for individual languages in Northern Eurasia. NEL comprises a list of $1016$ distinct \textit{concepts}\footnote{The concepts in NEL are given in German (see Table \ref{tab:concepts_domains}).} together with their word forms in $107$ languages (Table \ref{tab:concepts_domains}). Rare cases where a concept does not have any realization in a given language are excluded for that language.

\paragraph{Models $\&$ Corpus.}
In the main paper, for static word embeddings we use fastText\footnote{\url{https://fasttext.cc/docs/en/unsupervised-tutorial.html}} $300$-dimension word embeddings, trained on Wikipedia using the skip-gram model \citep{Bojanowski16}. 
For contextualised word embeddings (CE's) we use mBERT\footnote{\url{https://huggingface.co/bert-base-multilingual-uncased}} ({\it bert-base-multilingual-uncased} model) $768$-dimension vectors  for the $16$ languages.
To extract sentences for {\sc SNC-cloud}, we use the Leipzig corpus.\footnote{\url{https://corpora.uni-leipzig.de/en?corpusId=deu_news_2021}}
For concepts, their translations and semantic domains we use the North Euralex (NEL) dataset \citep{dellert2020northeuralex}. 
Other models and datasets that we experiment with, together with full explanations of how we use each dataset are detailed in Appendix \S\ref{app:sec:exp_setup}. 


\section{Cross-lingual Lexicon Alignment}
\label{sec:cross-lingual_lexicon_alignment} 

In this section, we lay the groundwork for cross-lingual lexicon alignment, focusing on word-level semantic similarity between languages. We use local metrics \citep{hamilton2016cultural,thompson18,Thompson_Lupyan20}, extend them to novel ones and provide a framework for comparing word meanings across languages.\footnote{In \S\ref{sec:related_work} we provide a further explanation as to why we choose this set of metrics.}

\subsection{Computational Framework}
\label{subsec:comp_frame}

The computational framework we adopt in this paper, which we term Semantic Neighborhood Comparison (SNC), relies on the relationships between the nearest neighbors of translation equivalents to compare embeddings across different spaces. This approach has been used in various forms for both computational historical linguistics and lexical similarity tasks \cite{hamilton-etal-2016-diachronic,Thompson_Lupyan20,Beinborn20}.
We experiment with several variants of this approach, including one based on contextualized word embeddings, which is, to our knowledge, novel.

\paragraph{Notation.}
Let $\mathcal{C}$ be the set of concepts in the NEL dataset \citep[see \S\ref{sec:exp_setup}]{NEL19}. We adopt the notion of a concept from the lexical typology literature \citep[e.g.,][]{NEL19,clics3}\footnote{A language $L \in \Omega$ may or may not lexicalize a concept $c \in \mathcal{C}$, and may lexicalize several concepts with one word (colexification).}, and take it to mean a word sense defined independently of any specific language. 
Let $\Omega$ be a set of languages.
We denote the lexicon corresponding to $\mathcal{C}$ in a given language $L$ with $\mathcal{L}$, and note that $|\mathcal{L}| \leq |\mathcal{C}|$ for every language.
We assume that $\mathcal{C}$ is partitioned into domains, and denote the (non-overlapping) domains with $\mathcal{D}_1,\ldots,\mathcal{D}_m$.

Given a concept $c\in \mathcal{C}$, we denote its lexicalization (the word expressing that concept) in language $L$ with $r_L(c) \in \mathcal{L}$. A translation pair between languages $L_1$ and $L_2$ is a pair of words $(w_1,w_2) \in \mathcal{L}_1 \times \mathcal{L}_2$, such that there exists $c \in \mathcal{C}$ such that $r_{L_1}(c)=w_1$ and $r_{L_2}(c)=w_2$. 
For example, the concept {\sc song} gives rise to the English-French translation pair \textit{(song,chanson)}.
In principle, several translation pairs may correspond to a concept and language pair, but in the data we experiment with, this does not occur. 

For a given word $w$ in a given language $L$, we denote its embedding with $emb(w,L)$. We denote the embedding space corresponding to $L$ with $\ell$.

\subsection{Alignment Metrics}
\label{subsec:alignment_metrics}

Let $c \in \mathcal{C}$ be a concept and $w_1 = r_{L_1}(c) \in \mathcal{L}_1$, $w_2 = r_{L_2}(c) \in \mathcal{L}_2$ its lexicalizations, and $v_1=emb(w_1,L_1) \in {\ell}_1$, $v_2=emb(w_2,L_2) \in {\ell}_2$ their respective embeddings. We compute  its $k$ nearest neighbors in ${\ell}_1$ with $\{{n_1^{(1)}},...,{n_k^{(1)}} \}$ ($k=100$).\footnote{See Appendix \S\ref{app:sec:exp_setup} for hyperparameters details.} We then translate the nearest neighbors to $L_2$ using the NEL dataset\footnote{Translation retrievel method explained in Appendix \S\ref{app:sec:exp_setup}.}, by taking their translation pairs, and denote the resulting vectors with $\{n_1^{(2)},...,n_k^{(2)} \} \in {\ell}_2$.

\paragraph{Neighbors Overlap ({\sc NO}).} 
A na\"{i}ve approach to comparing the meaning of a concept across languages is to compare the number of overlapping nearest neighbors of a word and its direct translation across languages \cite{thompson18}. 
This approach is intuitive and stems from the distributional definition of meaning as the semantic neighborhood of the concept. 

For a concept $c$, we back-translate its $k$ nearest neighbors in ${\ell}_1$ and ${\ell}_2$ to $\mathcal{C}$\footnote{To enable the intersection computation, the concepts need to reside in a ``joint space'', here, the concept space $\mathcal{C}$ can be thought of as an interlingua.} and define the alignment to be the amount of overlapping neighbors (in $\mathcal{C}$) divided by $k$.

\paragraph{Semantic Neighborhood Comparison (SNC).}

Although neighbors overlap has proven valuable for evaluating word-level similarities \cite{thompson18}, it falls short in capturing the intricate semantic relationships within the groups of neighbors. To address this limitation, we define the key metric in this paper, which serves as the foundation for all other variants. We define the unidirectional metric as

\begin{small}
\vspace{-.7cm}
\begin{multline*}
    a_{L_1 \to L_2}= \\ \rho\left(\left(cos(v_1,n_i^{(1)})\right)_{i=1}^k,\left(cos(v_2,n_i^{(2)})\right)_{i=1}^k\right)
\end{multline*}
\end{small}

$\rho$ is the Pearson correlation coefficient.\footnote{We conducted experiments with Spearman correlation, as well as Kendall $\tau$. They present similar trends and are omitted due to space considerations.} The bidirectional metric as the arithmetic mean over the two directions:

\begin{equation*}
    a_{L_1 \leftrightarrow L_2} = \frac{a_{L_1 \to L_2} + a_{L_2 \to L_1}}{2}
\end{equation*}

We refer to this alignment strategy as {\sc SNC-static}.


\paragraph{Contextualised Word Embeddings.} 

We now turn to detailing metrics that are analogous to {\sc SNC-static}, but instead use CEs.\footnote{We denote contextualised word embeddings by CEs.}

\paragraph{{\sc SNC-ave}} 
For word $w \in \mathcal{L}$, we extract its representation from all layers (if $w$ is tokenized to multiple subwords, we average over the subword representations). 
We average the outputs from layers $1$-$12$ to define the final vector for $w$. We then proceed with the SNC process, as described with {\sc SNC-static}.\footnote{We follow the work of \citep{vulic_probing_lex_20} on probing contextualized models for lexical semantics, averaging across layers bottom-to-top. However, we experimented with alternative settings, such as pooling from the top layer or averaging up to layer n (n<12), and found this method to be the most stable overall.}

\paragraph{{\sc SNC-cloud}} For word $w \in \mathcal{L}$, we extract all sentences (with a threshold of $1000$) that $w$ appears in, from an auxiliary corpus (see \S\ref{sec:exp_setup}). We extract the CEs (from layer $12$, if it is tokenized to subwords, we average over them) for $w$ from each of the sentences. Denote these vectors with $V_w = \{v_{1_w},...,v_{k_w}  \} \subseteq \mathbb{R}^{768}$. 
In this setting, each word $w$ is represented by a point cloud of vectors $V_w$. Hence, the distance between two words is the distance between their corresponding point clouds. We  define \textit{point-cloud distance} as follows: 

\begin{equation*}
    d(w,\tilde{w}) = min_{i,j}~cos({v_{i_{w}}},{v_{j_{\tilde{w}}}})
\end{equation*}

We follow the {\sc SNC} procedure (defined above) under this definition of distance.

\section{Validation Experiments for {\sc SNC}}
\label{sec:validation}

Due to the opaque nature of distributional metrics, it is important to verify their validity as metrics for cross-lingual alignment of translation pairs. While some of the metrics we use are taken from the literature, others have not yet been used for lexical alignment , as they were either adapted from other tasks of lexical similarity in NLP, or are novel adaptations of existing metrics.
Unlike in many NLP papers and tasks, there is no gold standard definition of the extent to which the meanings of two words in different languages align, which prohibits direct validation of the metrics \cite{schlechtweg-etal-2020-semeval}.
Instead, we conduct two sets of experiments (synthetic and naturalistic) to verify both the self-consistency of the metrics (Sec \S\ref{subsec:validation_synthetic}) and its validity against a high quality reference (\S\ref{subsec:validation_lex_gaps}).


\begin{figure}
    \centering
    \includegraphics[width=\columnwidth]
    {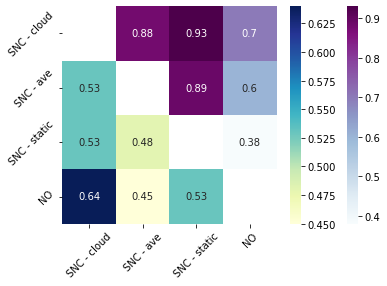}
    \caption{Correlation between the various metrics. Pearson correlation is computed for different aggregation methods. The \textbf{upper} matrix represents concept-level correlations, while the \textbf{bottom} matrix represents domain-level correlations. All correlation values are significant with $p<0.05$. }
    \label{fig:heatmap_measures}
\end{figure}

\subsection{Synthetic Validation} 
\label{subsec:validation_synthetic}

Given the lack of human-evaluated data quantifying the similarity between translation equivalents across languages, we conduct synthetic experiments to verify the metrics' internal consistency. These experiments evaluate the proposed metrics without relying on any external resources. 

\paragraph{Shuffle Baseline.}
To verify that the degree of alignment of words (or semantic domains) is a product of their similar semantic structure with neighboring words, rather than some unexpected artifact, we conduct a shuffle baseline test. This test evaluates whether the observed degree of alignment is a result of words being in a dense/sparse part of the embedding space, which may skew the results. To compute the shuffled alignment, let $\{{n_1^{(1)}},...,{n_k^{(1)}} \}$ be the $k$ nearest neighbors of $w_1 \in L_1$, and $\{{n_1^{(2)}},...,{n_k^{(2)}} \}$ their translation equivalents in $L_2$. We perform a random permutation of the translations indices, such that, $\{ {n_1^{(2)}},...,{n_k^{(2)}} \} \mapsto \{ {n_{p(1)}^{(2)}},...,{n_{p(k)}^{(2)}} \}$, with $p$ a permutation over $\{1,...,k\}$. We then compute $a_{L_1 \to L_2}$ (as defined in \S\ref{subsec:alignment_metrics}), and do the same for the other direction $a_{L_2 \to L_1}$). 
If a high alignment is an artifact of the neighbors being in a dense/sparse part of the embedding space, the correlation between the alignment and its shuffled version should be close to $1$ (and to $0$ if it is not the case). We find that the correlation is always $r \in [-0.5,0.5]$ with $p$-values $p>0.5$, suggesting that the density does not play a major role in the observed correlations.

\paragraph{Sensitivity.}
A key step in {\sc SNC} computation (Section  \S\ref{subsec:alignment_metrics}) is the $k$ nearest neighbors search, which is restricted to the NEL lexicon. We verify that results are robust to removal of semantic domains from the data by removing $j$ domains ($j=5,10,15$) and computing the correlation between the results before\footnote{The results before are the results aggregated by domain (see \S\ref{subsec:dom_level_comp}), computed on the full domain list, and then the domains that are selected to be removed are removed from the vector before making the comparison.} and after the removal (we do this $1000$ times per $j$). The results are highly stable to such removal, with $0.87 \leq r \leq 0.99$ and $p \leq 0.05$.\footnote{This holds true for all {\sc SNC} variants, models and datasets we experiment in the paper.}

\subsection{Naturalistic Validation}
\label{subsec:validation_lex_gaps}
 
In this section, we present a novel external validation method for cross-lingual lexicon alignment, using a recently developed, extensive resource that identifies lexical gaps across languages.
A key notion in capturing lexical diversity across languages is that of the \textit{lexical gap}, which refers to the lack of lexicalization of a particular concept in a particular language. For example, many languages lack an equivalent of the English word \textit{cousin}, and instead employ several more specific words that distinguish male and female, elder and younger or paternal and maternal cousins \cite{Khishigsuren23}. We use a newly released lexical resource for the kinship domain, which contains 37370 gaps in 699 languages.\footnote{\url{https://github.com/kbatsuren/KinDiv}} The resource focuses on the domain of kinship as it is universally represented in human languages, but is also known to be incredibly diverse across languages and cultures. This is a unique linguistic resource, as it is the first extensive resource to cover lexical gaps across a diverse set of languages. As such, it allows for a reliable external validation of the alignment methods discussed in this paper.  

\begin{figure*}[tbh]  
      \centering
      \includegraphics[width=1\linewidth]
    {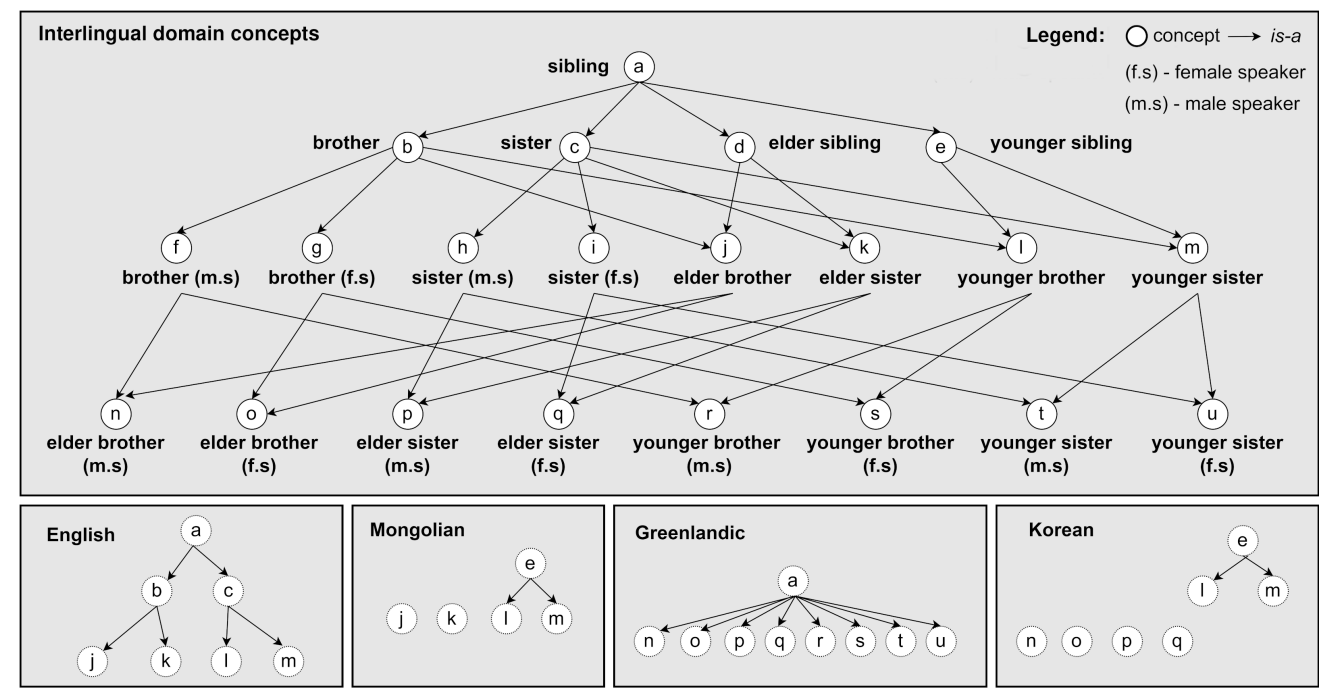}
    \caption{Lexical Gaps. Interlingual conceptual layer of sibling domain, reproduced with permission from the authors of \citep{Khishigsuren23}.}
    \label{fig:lex_gaps}
\end{figure*}

We consider similar gap patterns as indicative of greater alignment between languages. Keeping this in view, we establish a metric for alignment derived from these gap patterns, which we view as a high quality reference point.

\paragraph{Lexical Gaps.} The notion of a lexical gap is closely related to that of untranslatability \cite{catford78}.  
For example, \citet{Wierzbicka2008_color} considered that the concept of {\it color} is a lexical gap in Warlpiri, an Australian Indigenous language, as it lacks a word for it. A lexical gap is defined as the absence of a specific word or term in a language to express a particular concept or idea \cite{Bentivogli}.

\paragraph{Interlingua.} 

In order to compare lexical gaps across languages, an interlingual conceptual space of the kinship domain is defined (Figure \ref{fig:lex_gaps}). It consists of 198 concepts and 347 is-a relations (e.g., \textit{parent's male sibling}), covering the six subdomains that Kinship is usually divided into: grandparents, grandchildren, siblings, uncles and aunts, nephews and nieces, and cousins \cite{Murdock1970}. 
We denote the subdomains by $\mathcal{S}$. For subdomain $s \in \mathcal{S}$ (e.g, \textit{sibling}), let $\mathcal{C}_s$ be the list of concepts associated with it (i.e, all the possible lexicalizations in the interlingual space, e.g, \textit{female elder sister}).
For a language $L$ and subdomain $s \in \mathcal{S}$, $c \in \mathcal{C}_s$ is a \textit{gap} if it is not distinctly lexicalized in $L$. We define the \textbf{gap pattern} of subdomain $s$ as the set of lexical gaps for this concept, and denote it with $\xi_{L,s}$.

\paragraph{Comparing Lexical Gaps.} 

To use the information about the lexical gaps as validation, we define a metric for alignment of language pairs, based on their gap patterns. 
Let $L_1, L_2 \in \Omega$ be a pair of languages, we define: 

\begin{equation*}
    \lambda_{L_1 \leftrightarrow L_2} = \frac{1}{|\mathcal{S}|}{\Sigma}_{s \in \mathcal{S}}\frac{|\xi_{L_1,s} \cap \xi_{L_2,s}|}{|\mathcal{C}_s|} 
\end{equation*}

We denote the concatenation over all pairs of languages in ${\Omega}\choose{2}$ with $\lambda$.
To compare the above alignment with {\sc SNC} and {\sc ColexA}, we manually filter the concepts $\mathcal{C}$ that {\sc SNC} and {\sc ColexA} is computed on (i.e, $\mathcal{C}$) to contain only relevant concepts that are both in the Kinship domain and that can be manually mapped into a subdomain $s \in \mathcal{S}$ (e.g, the concept ``onkel::N'' for {\it uncle} in $\mathcal{C}$ is mapped to the subdomain \textit{uncles and aunts}, however the concept ``ihr::PRN'' for {\it you} cannot be mapped to any subdomain $s \in \mathcal{S}$). We denote the filtered concepts with $\tilde{\mathcal{C}}$ and restrict the computation of {\sc SNC} and {\sc ColexA} to $\tilde{\mathcal{C}}$. 
We compute correlation  at the word-level (the domain-level is not relevant here, as we are restricted to one domain, kinship) between the various metrics as detailed in \S\ref{subsec:word_level_comp}. For each alignment measure and pair of languages $(L_j,L_p) \in {{\Omega}\choose{2}}$, we have a vector in $\mathbb{R}^{|\tilde{\mathcal{C}}|}$. We perform aggregation over its components which results in $\mu_{L_j,L_p} \in \mathbb{R}$. The final vector $\mu \in \mathbb{R}^{{\Omega}\choose{2}}$ is a concatenation over all pairs of languages. For each alignment measure we compute the Pearson correlation between $\mu$ and $\lambda$.


\section{Analysis}
\label{sec:colex_dista_comp}

Having established a set of metrics and an extensive validation suite, we now turn to analyzing the alignment results. This analysis is conducted at two levels of granularity: the word level and the domain level. We compare various metrics, identifying which words and domains are most and least aligned. Furthermore, we examine factors affecting alignment, such as lexical properties and environmental features, to uncover the underlying causes of semantic divergence across languages.

\quash{
Let $\mathcal{M}$ be the set of alignment metrics. We denote the raw data as follows:

\vspace{-.6cm}
\begin{equation*}
\mu(m,L_p,L_j) \quad\forall~m \in \mathcal{M}, L_p \times L_j \in \Omega^2
\end{equation*}

For a pair of languages $L_p$, $L_j$ and a metric $m$, $\mu(m,L_p,L_j) \in \mathbb{R}^{|\mathcal{C}|}$ is a vector whose $i$-th coordinate is the alignment value of concept $c_i$ under metric $m$ between $L_p$ and $L_j$.

Throughout the following section we use Pearson's $r$ (with a two-tailed test for significance) for computing correlation, unless stated otherwise.

}


\begin{table}[!t]
\centering
\footnotesize
\renewcommand{\tabcolsep}{0.15cm}
\begin{tabular}{cccc}
\toprule

&
{\sc SNC-static}& 
{\sc SNC-ave} &
{\sc SNC-cloud} \\

\midrule

\multirow{3}{*}{\rotatebox[origin=c]{90}{Top 3}} & March & twelve  & thirty \\
& August & eleven  & fifty \\
& January & five & twelve \\

\midrule

\multirow{3}{*}{\rotatebox[origin=c]{90}{Bottom 3}} & rise & be afraid & corner \\
& groan & soft & soft \\
& set & be noisy & round \\

\bottomrule
\end{tabular}
\caption{Most and least aligned words. Word-level alignment, averaged across languages.}
\label{tab:topbottom_words}
\end{table}


\subsection{Word-level $\&$ Domain-level Comparison}

\label{subsec:word_level_comp}

\paragraph{Word-level Comparison.} We start with a comparison between the metrics themselves. 
The most straightforward level of comparison between metrics is their word-level correlation\footnote{Each metric, a concept and language pair, give rise to a vector of alignment scores (full details in Appendix \S\ref{app:subsec:word_level_corr}).}.

Figure \ref{fig:heatmap_measures} presents the Pearson correlation between the metrics, and Table \ref{tab:topbottom_words} shows the top/bottom aligned words for the metrics. 
Results show that {\sc SNC} methods are moderately correlated among themselves ($r$ around $0.5$), meaning there is a substantial variability in their predictions at the word level.
Moreover, manual inspection of the data reveals that it is challenging to infer conclusions at the word level off handily (Section \S\ref{sec:qualitative}). 


\paragraph{Domain-level Comparison}
\label{subsec:dom_level_comp}

Alignment metrics between languages are often used to compare the degree of alignment across different domains. For example, \citet{Thompson_Lupyan20} argue, based on findings with a {\sc SNC-static} metric, that more structured domains (e.g, Quantity) tend to be better aligned across languages. 
To examine the alignment at the domain level, for every measure $m \in \mathcal{M}$, we aggregate the word-level alignment over each domain (without aggregating over languages). 
Strikingly, as opposed to the word-level comparison, here the similarity between the methods is very high, reaching $r=0.93$ (between {\sc SNC-cloud} and {\sc SNC-static}). 
This finding encourages the formulation of conclusions at the domain level, as it presents to be more stable.


\quash{
\begin{table}[!t]
\centering
\footnotesize
\renewcommand{\tabcolsep}{0.15cm}
\begin{tabular}{cccc}
\toprule

&
{\sc SNC-static}& 
{\sc SNC-static} &
{\sc SNC-cloud} \\

\midrule

\multirow{3}{*}{\rotatebox[origin=c]{90}{Top 3}} & Quantity & Quantity & Quantity \\
& Time & Time & Kinship \\
& Kinship & Kinship & Time \\

\midrule

\multirow{3}{*}{\rotatebox[origin=c]{90}{Bottom 3}} & Possession  & Basic actions  & Agriculture \\
& Basic Actions & Motion & Spatial relations \\
& Motion & The house & The house \\

\bottomrule
\end{tabular}
\caption{Most and least aligned domains for various metrics. Alignment computed by aggregating over languages and over domains. ``Basic actions.'' refers to ``Basic actions and technology'' and ``Agriculture'' refers to ``Agriculture and vegetation''} 
\label{tab:topbottom_domains}
\end{table}
}

\begin{figure}
    \centering
    \includegraphics[width=\columnwidth]{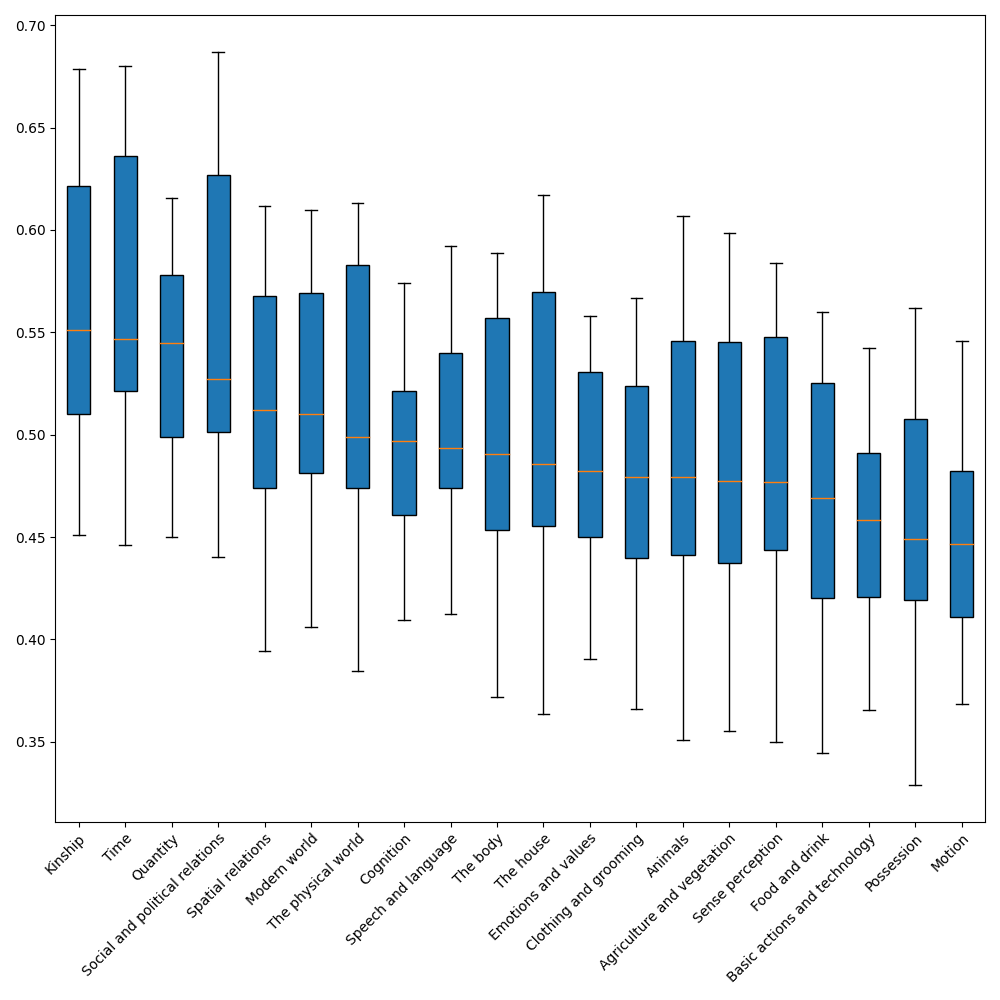}
    \caption{Alignment of domains under {\sc SNC-ave}. The domains are ranked according to the mean value of the alignment. Each box represents the distribution of alignment values (per language pair), for a specific domain (concepts-level alignment is aggregated within each domain). The centre line is the median, the box limits are the upper and lower quartiles, and the whiskers represents the $1.5 \times$ interquartile range.}
    \label{fig:box_plot_bert_cloud}
\end{figure}

\paragraph{Relative Degree of Alignment across Domains.} We examine what domains are the most/least aligned across languages. Figure \ref{fig:box_plot_bert_cloud} shows both the distribution and median of alignment values for each language pair across the semantic domains, for {\sc SNC-cloud}. The most aligned domains are Quantity, Time and Kinship\footnote{This trend persists for all {\sc SNC} methods, across all model architectures, and various $k$ values.}, whereas the least aligned domains are Motion, Basic Actions, and Technology and Possession. Similar trends are reported by \citet{Thompson_Lupyan20}, who argue that the high degree of alignment of these domains is related to their structure and organization along explicit dimensions (e.g., generation: grandmother/mother/daughter)
capture different notions of similarity. 
Table \ref{tab:topbottom_domains} presents a few examples of the differences.

\subsection{Determinants Of Semantic Similarity}
\label{subsec:lexical_env_feats}

In this section we examine the effect of such features on the measured correlations. A combination of lexical (frequency, concreteness, rate of change) and environmental (cultural and geographic distance) features was selected. See \S\ref{sec:exp_setup}.  

\paragraph{Correlation With Lexical Features.} 
At the word-level there is no correlation with respect to frequency and concreteness, and weak-moderate negative correlation with rate of lexical change. When aggregating over domains concreteness is still not correlated with any of the alignment methods; however, the correlation goes up for frequency (albeit still weakly for {\sc SNC}), and interestingly a substantial correlation for {\sc NO} ($r=0.6$). and jumps for rate of change ($r \approx -0.6$ for {\sc SNC}). This interesting result means that words that undergo faster lexical change are less aligned across languages.
This echoes the findings that polysemy has an important role in the rate of lexical change \cite{brown83,Thompson_Lupyan20}, and coincides with the findings that rate of change is correlated negatively with prototypicality (how representative a word is
of its category) \cite{dubossarsky-etal-2017-outta}. Our experiments demonstrate that \textbf{polysemy} also plays a role in lexical alignment, as further elaborated in Appendix \ref{app:polysemy}.


\begin{table}[!t]
\centering
\footnotesize
\renewcommand{\tabcolsep}{0.15cm}

\begin{tabular}{cp{0.8cm}p{0.9cm}p{0.9cm}p{0.9cm}p{0.9cm}}
\toprule
& &
{\sc SNC-cloud} &
{\sc SNC-ave} &
{\sc SNC-static} &
{\sc NO} \\ \midrule
\multirow{2}{*}{\textbf{CLT}}
& C & $0.14^{\star}$ & $0.1^{\star}$ & $0.25^{\star}$ & -$0.08$ \\
& D & $0.2^{\star}$ & $0.49^{\star}$ & $0.13^{\star}$ & $0.11^{\star}$ \\ \midrule
\multirow{2}{*}{\textbf{GEO}}
& C & $0.03^{\star}$ & $0.09^{\star}$ & $0.22^{\star}$ & -$0.05^{\star}$ \\
& D & $0.16^{\star}$ & $0.41^{\star}$ & $0.05$ & $0.1^{\star}$ \\ \midrule
\multirow{2}{*}{\textbf{frequency}}
& C & $0.04^{\star}$ & $0.06$ & $0.06$ & $0.1^{\star}$ \\
& D & $0.33^{\star}$ & $0.18^{\star}$ & $0$ & $0.6^{\star}$ \\ \midrule
\multirow{2}{*}{\textbf{concreteness}}
& C & $0.03$ & $0$ & $0$ & $0$ \\
& D & $0.18^{\star}$ & $0.06$ & $0.1^{\star}$ & $0.1$ \\ \midrule
\multirow{2}{*}{\textbf{rate-change}}
& C & -$0.32^{\star}$ & -$0.22^{\star}$ & -$0.25^{\star}$ & -$0.14^{\star}$ \\
& D & -$0.57^{\star}$ & -$0.62^{\star}$ & -$0.62^{\star}$ & -$0.42^{\star}$ \\
\bottomrule
\end{tabular}

\caption{Correlation with lexical and enviromental features. Columns represent the features (CLT denotes cultural distance and GEO denotes geographical distance) and subcolumns represents concept-level aggregation (C) vs. domain-level aggregation (D). {\sc NO} represents Neighbors Overlap metric. significant correlation with $p < 0.05$ are marked by $\star$. }
\label{tab:corr_lexical_enviro}
\end{table}

\paragraph{Correlation With Environmental Features.} 
The question of how \textbf{geographical} and \textbf{cultural} factors influence the alignment of words across languages is a matter of ongoing discussion among scholars \cite[e.g.,]{youn2016,Josserand21}.
Table \ref{tab:corr_lexical_enviro} shows a significant correlation with geographic and cultural distance for {\sc SNC}, with cultural distance playing a more prominent role.


\section{Qualitative Analysis}
\label{sec:qualitative}

We conduct a small scale qualitative analysis to further understand the results.\footnote{The qualitative analysis was conducted by one of the authors. The full details of the setup are in Appendix \S\ref{app:sec:qualitative}.}

Drawing conclusions at the word level solely from the raw data is challenging, as evidenced by the discrepancies between methodologies in our empirical analysis. We give a few examples of factors that impact the alignment.\footnote{Analysis done for {\sc SNC-cloud}.} The first is cultural differences. To illustrate, the word ``pig'' is less aligned between English and Arabic than between English and German. Manual examination of the data reveals that the nearest neighbors of ``pig'' in English and German include animals and food items, such as ``butter'' and ``salt,'' whereas in Arabic, they are exclusively animals. 
This difference might arise from the cultural context in Islam, where pork is prohibited. Another reason for divergence is illustrated by the word ``soft,'' which is one of the least aligned words across languages. The neighbors of ``soft'' are highly varied and come from different semantic domains, even within the same language, making their connection to the word itself less immediate. For example, in English, some neighbors for the word ``soft'' are ``red'', ``sea'' and ``hand,'' while in German, they include ``meat,'' ``beautiful'' and ``rich''. This diversity in semantic association contributes to the lower alignment for words like ``soft''. 

Differences in word senses also contribute to misalignment. For example, the word ``brother'' is less aligned between English and Hebrew, which might be attributed to the homonymy in Hebrew between the senses of ``brother'' and ``hearth''. This difference is reflected in their nearest neighbors. This issue is directly related to the influence of polysemy on alignment, which we address in Appendix \S\ref{app:polysemy} and leave to future work.

\section{Discussion}
\label{sec:discussion}

How and why languages vary in their use of words to carve up the semantic space has long been a question of central interest in cognitive sciences and linguistics, but has often been overlooked in NLP. Recent advances in NLP allow addressing this gap, and performing such analyses well and at scale.  In this paper, we formulate this question as an NLP task, and provide a methodology for computing the efficacy of different metrics in addressing the task. We use existing metrics and extend them to contextualized word representations. We evaluate the metrics across multiple scenarios, using both synthetic and naturalistic validation approaches.
We observe consistent trends across all metrics and architectures: the rate of change is a strong predictor of alignability. Additionally, internally structured domains such as Time, Quantity, and Kinship show the highest degree of alignment across languages, consistent with \citep{Thompson_Lupyan20}.

One of the major challenges in analyzing cross-lingual alignment for individual words or domains is the lack of ground-truth data for validation. This impedes the comparison between different metrics, and thereby hinder progress on this task.
To address this, we provide both synthetic and naturalistic validations using a newly created linguistic resource, based on the kinship domain. Our validation shows that all the {\sc SNC} metrics we propose effectively capture cross-lingual semantic variability, with a slight preference for the metrics that are based on contextualized representations.

In future work, we plan to leverage cross-lingual alignment to investigate its impact on cross-lingual transfer on various NLP tasks. Additionally, we aim to use this approach to examine the extent to which LLMs encode cultural knowledge, a topic that has recently garnered significant attention \citep{havaldar2023multilingual,li2024culturellm,rao2024normad,zhou2024does}.


\section*{Limitations}

While we introduce a novel validation for the task of cross-linguistic lexicon alignment, it is currently limited to the kinship domain. A more comprehensive validation spanning multiple domains would provide a more thorough verification of the metrics. Additionally, we use the NEL dataset to analyze 1,016 concepts across various languages. In future work, we plan to expand this list to include more words and languages, enhancing the robustness of our analysis.

Moreover, throughout our study, we use the semantic domains defined in the NEL dataset. Although these manually defined domains are widely used in cognitive science and linguistic research, manual examination has shown that they are not optimal, due to noise and what seems to be inconsistent rationale as to what to include and what not. They could benefit from additional filtration and refinement to improve their accuracy and relevance.

Our work is readily applicable to both static word representations and contextualized representations; however, it is not currently suited for autoregressive models such as GPT and its variants. In future work, we aim to expand our metrics and evaluation to accommodate these architectures as well.


\bibliography{anthology, custom}
\bibliographystyle{acl_natbib}

\appendix

\section{Experimental Setup} \label{app:sec:exp_setup}

In this section we provide further details regarding our experimental setup.

\paragraph{Languages.} We perform our analysis on a diverse set of $16$ languages, spanning 7 different top-level language families from many geographical areas across Eurasia: English (eng), French (fra), Italian (ita), German (deu), Dutch (nld), Spanish (spa), Polish (pol), Finnish (fin), Estonian (est), Turkish (tur), Chinese (chn), Korean (kor), Japanese (jap), Hebrew (heb), Hindi (hin) and Arabic (arb). 

\paragraph{NorthEuraLex}
(\textbf{NEL}) is a lexical resource compiled from dictionaries and other linguistic resources available for individual languages in Northern Eurasia \citep{dellert2020northeuralex}. NEL comprises a list of $1016$ distinct \textit{concepts}\footnote{The concept in NEL are given in German.} together with their word forms in $107$ languages (Table \ref{tab:concepts_domains}). 

Rare cases where a concept does not have any realization in a given language are excluded for that language.

\paragraph{Semantic Domains.}
We map the concepts in NEL to domains, using Concepticon.\footnote{\url{https://concepticon.clld.org/}}
There are $20$ domains, each containing $22-136$ concepts (number of concepts is written next to each domain): animals (47), agriculture and vegetation (23), time (68), quantity (40), kinship (26), basic actions and technology (140), clothing and grooming (27), cognition (30), emotions and values (54), food and drink (42), modern world (28), motion (70), possession (26), sense perception (50), social and political relations (30), spatial relations (85), speech and language (25), the body (94), the house (20) and the physical world (75).

\paragraph{Word Embeddings.} 
In the main paper, for static word embeddings we use fastText\footnote{\url{https://fasttext.cc/docs/en/unsupervised-tutorial.html}} $300$-dimension word embeddings, trained on Wikipedia using the skip-gram model \citep{Bojanowski16}. For contextualised word embeddings (CWE) we use mBERT\footnote{\url{https://huggingface.co/bert-base-multilingual-uncased}} ({\it bert-base-multilingual-uncased} model) $768$-dimension vectors  for the $16$ languages.
To extract sentences for {\sc SNC-cloud}, we use the Leipzig corpus.\footnote{\url{https://corpora.uni-leipzig.de/en?corpusId=deu_news_2021}} 
Due to lack of space we choose to focus on this setup after experimentation with other models architectures as trends are consistent across the board.
We also run our experiments also on XLM-RoBERTa-base \footnote{\url{https://huggingface.co/xlm-roberta-base}} for {\sc SNC-cloud} and {\sc SNC-ave} and on 300-dim word2vec multilingual embeddings \footnote{\url{https://github.com/Kyubyong/wordvectors}} for {\sc SNC-static}. Furthermore, we perform all computations for {\sc SNC-cloud} and {\sc SNC-ave} with a different dataset; the Wikipedia section in the Leipzig Corpus, for the latest year available in each language \footnote{\url{https://wortschatz.uni-leipzig.de/en}}. The trends were highly similar to what we report in the paper.\footnote{See Appendix \ref{app:sec:other_arch} for experiments on other architectures than the ones presented in the main paper.}.

\paragraph{Hyperparameters.} For our distributional based alignments {\sc SNC} and {\sc NO} (\S\ref{subsec:alignment_metrics}), we set $k=100$. We experimented with other values of $k$ ($k=10,50,1000$) and selected the one that overall correlated the most with our validation and showed more robust results in terms of correlation with other features (such as lexical ans enviromental features). This is also the hyperparameter chosen in the original work of \citep{Thompson_Lupyan20} for the {\sc SNC-static} methods, based on similar reasons.

\paragraph{Lexical and Language Features.} 
We report results while controlling for a variety of lexical features and features of the languages compared.
Geographic distance between languages is computed as the geodesic distance (distance in an ellipsoid) between their latitude and longitude coordinates (taken from Glottolog\footnote{\url{https://glottolog.org/}}). 
Cultural distance is computed as the proportion of common cultural traits from a set of 92 non-linguistic cultural traits for $16$ societies representing the languages in our analysis, taken from D-PLACE\footnote{\url{https://d-place.org/}} \cite{Thompson_Lupyan20}.
We use the {\it wordfreq} library\footnote{\url{https://pypi.org/project/wordfreq}} for word frequencies. We then compute the log-transformed frequency (to reduce the impact of outliers and extreme values).  

\paragraph{Rate of Lexical Change.} 
Realizations of some concepts, such as {\it tail},  evolve rapidly, while others, such as {\it two} evolve at a much slower rate. This phenomenon is referred to as the {\it rate of (lexical) change}. We use lexical change rates derived from \citep{Pagel07}, available for Russian, Greek, English and Spanish. 

\begin{figure}
    \centering
    \includegraphics[width=\columnwidth]
    {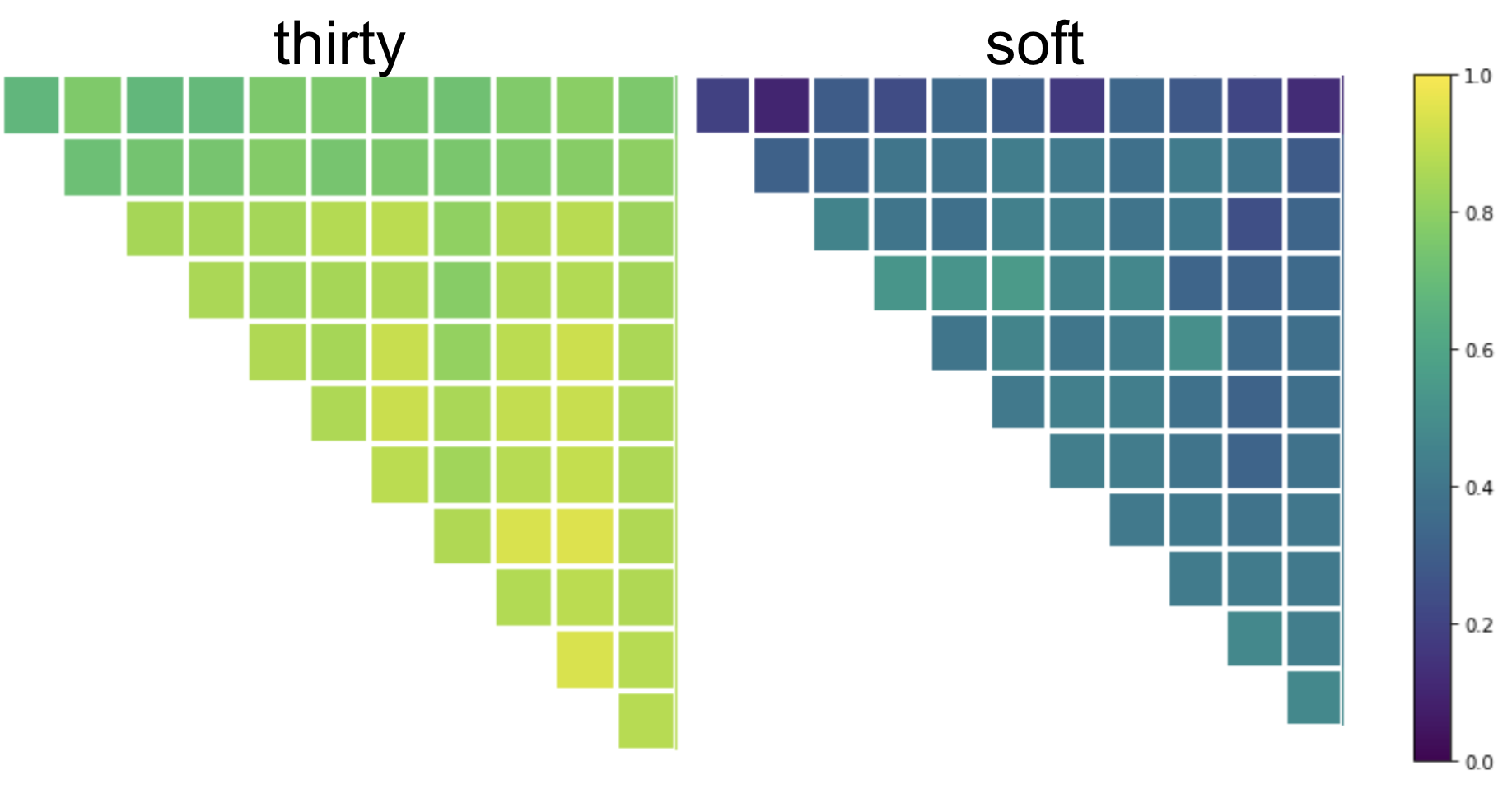}
    \caption{Alignment of individual concepts under {\sc SNC-cloud}. Left: alignment for the concept ``thirty''. Right: alignment for the concept ``soft''. Computed for a sample of $12$ languages. Each square represent alignment for a language pair. More aligned pairs are closer to $1$. The alignment scores are computed for: Turkish, Hebrew, German, Finnish, English, Dutch, Spanish, Russian, French, Hindi, Italian, and Arabic. }
    \label{app:fig:heatmap_concepts_high_low}
\end{figure}

\section{Word-level and Domain-level Analysis}
\label{app:word_level_dom_level_analysis}

We hereby describe in full details how we perform the analysis at the word-level and the domain-level. 
Let $\mathcal{M}$ be the set of alignment metrics. We denote the raw data as follows:

\vspace{-.6cm}
\begin{equation*}
\mu(m,L_p,L_j) \quad\forall~m \in \mathcal{M}, L_p \times L_j \in \Omega^2
\end{equation*}

For a pair of languages $L_p$, $L_j$ and a metric $m$, $\mu(m,L_p,L_j) \in \mathbb{R}^{|\mathcal{C}|}$ is a vector whose $i$-th coordinate is the alignment value of concept $c_i$ under metric $m$ between $L_p$ and $L_j$.

Throughout the following section we use Pearson's $r$ (with a two-tailed test for significance) for computing correlation, unless stated otherwise.

\subsection{Word-level Correlations.}
\label{app:subsec:word_level_corr}

The most straightforward level of comparison between metrics is their word-level correlation. Let  ${\Omega}\choose{2}$ be the set of all language pairs (without repetitions), and denote its size with $l$=${|\Omega|}\choose{2}$. 
For $m \in \mathcal{M}$, define $\hat{\mu}(m) \in \mathbb{R}^{l|\mathcal{C}|}$ the concatenation of $\mu(m,L_p,L_j)$ 
for all language pairs. Word-level correlation is the Pearson correlation between $\hat{\mu}(m)$, for $m \in \mathcal{M}$ (See Figure \ref{app:fig:heatmap_measures} and Figure \ref{app:fig:heatmap_measures}). 

\subsection{Domain-level Correlations.}
\label{app:subsec:word_level_corr}

Alignment metrics between languages are often used to compare the degree of alignment across different domains. For example, \citet{Thompson_Lupyan20} argue, based on findings with {\sc SNC-static} , that more structured domains tend to be better aligned across languages. 
To examine the alignment at the domain level, for every measure $m \in \mathcal{M}$, we aggregate the word=level alignment over each domain (without aggregating over languages). We get $\hat{\mu}(m) \in \mathbb{R}^{lm}$ ($m$ is the number of semantic domains).

\begin{table}[!t]
\centering
\footnotesize
\renewcommand{\tabcolsep}{0.15cm}
\begin{tabular}{cccc}
\toprule

&
{\sc SNC-static}& 
{\sc SNC-static} &
{\sc SNC-cloud} \\

\midrule

\multirow{3}{*}{\rotatebox[origin=c]{90}{Top 3}} & Quantity & Quantity & Quantity \\
& Time & Time & Kinship \\
& Kinship & Kinship & Time \\

\midrule

\multirow{3}{*}{\rotatebox[origin=c]{90}{Bottom 3}} & Possession  & Basic actions  & Agriculture \\
& Basic Actions & Motion & Spatial relations \\
& Motion & The house & The house \\

\bottomrule
\end{tabular}
\caption{Most and least aligned domains for various metrics. Alignment computed by aggregating over languages and over domains. ``Basic actions.'' refers to ``Basic actions and technology'' and ``Agriculture'' refers to ``Agriculture and vegetation''} 
\label{tab:topbottom_domains}
\end{table}


\begin{figure}
    \centering
    \includegraphics[width=\columnwidth]
    {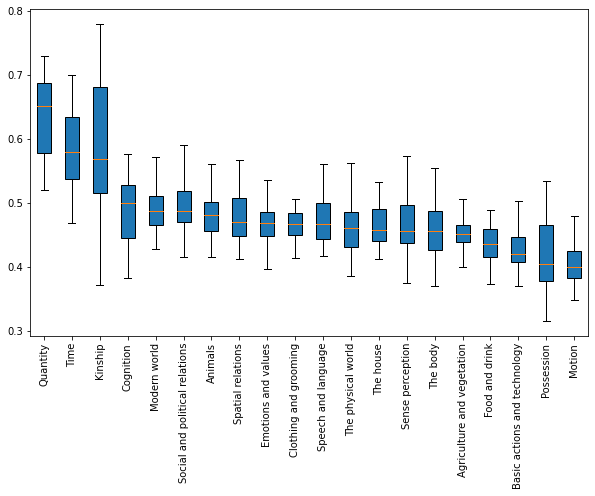}
    \caption{Alignment of domains under {\sc SNC-cloud}. The domains are ranked according to the mean value of the alignment. Each box represents the distribution of alignment values (per language pair), for a specific domain (concepts-level alignment is aggregated within each domain). The centre line is the median, the box limits are the upper and lower quartiles, and the whiskers represents the $1.5 \times$ interquartile range. Alignment computed using XLM-RoBERTa-base.}
    \label{app:fig:box_plot_rob}
\end{figure}

\section{Controlling for Lexical and Enviromental Features.}
To further examine the influence of lexical and environmental features on the alignment methods, we perform partial correlation tests to control for the various features, and multiple regression analysis to account for the overall variance that is explained by them. We compute the partial correlation\footnote{For the partial correlation computations we use the {\it pingouin} package \url{https://pingouin-stats.org/build/html/generated/pingouin.partial_corr.html}} between {\sc SNC} methods,  while controlling for the lexical and environmental features. 

We find that at the word-level measures are still moderately correlated with $r \approx 0.4$. 
At the domain-level, the methods are still highly correlated with one another ($r \approx 0.9$).
We use multiple linear regression to compute the adjusted $R$-squared value, with the environmental and lexical features as response variables. 
While the features explain $\approx 20 \%$ of the variance for {\sc SNC}. However, when aggregating over domains, the features explain up to $44\%$ of the variance for {\sc SNC}. This suggests that the analysis is more suitable at the domain-level.  


\section{Comparing {\sc SNC} to Norm-based Approaches}
\label{sec:norm_based_app}

In order to capture cross-lingual lexical similarity, cognitive scientists have used behavioral stimuli, e.g., sets of pictures that are named by speakers of different languages \cite{Thompson_Lupyan20}. Where the same pictures are named consistently in two languages with a given pair of words, these words are interpreted as semantically similar \cite{glaser92,jacobi22}. Another common paradigm is the use of translation norms \cite{hermans96}. In this section we discuss two such datasets, and empirically compare how well they align with {\sc SNC}. We note that these measures do not consider as ``ground truth'' but rather as converging evidence for the validity of the metrics, as it is not trivial that they measure the same (or even higly similar) things. 
We use two manual external resources: name-agreement in picture naming  \cite[Multipic;][]{multipic2018} and English-Dutch translation norms \cite[TransSim;][]{Tokowicz2002}. 
MultiPic is a standardized set of 750 drawings of concrete objects with name agreement norms for six European languages (English, Spanish, Netherlands Dutch, German, French and Italian). For each picture and language, the norm is an information statistic that reflects the level of agreement across participants. 
TransSim is a dataset of 562 Dutch-English translation pairs together with a human similarity rating between each pair.

\paragraph{Multipic.} 

We filter the pictures in the Multipic dataset to only include pictures with concepts from NEL. This results in a total of 194 pictures.
We compute the correlation between the agreement scores (average agreement score over all languages) for these  pictures and the different {\sc SNC}. 
We get that while {\sc SNC-ave} and {\sc SNC-static} are moderately correlated with Multipic ($r \approx 0.3$, $p<0.05$), the other methods are weakly to not correlated with the dataset.

\paragraph{TransSim.} 

We again filter the dataset to include word pairs that are covered by NEL, resulting in 187 Dutch–English translation similarity judgment scores. 
We compute the correlation between English-Dutch translation similarity judgements and the alignment metrics for English-Dutch, aggregated by domain (domain-level). A relatively high correlation is presented, where {\sc SNC-static} ($r=0.59$, $p<0.05$) and  {\sc SNC-ave} ($r=0.51$, $p<0.05$) rank highest.

To conclude, we find that overall {\sc SNC} are moderately correlated with the human-evaluated datasets, which may reflect a relative similarity in the notion of alignment captured by these datasets and the distributional metrics. We defer a more elaborate multi-approach comparison to future work.


\section{Qualitative Analysis}
\label{app:sec:qualitative}

We conduct a small scale qualitative analysis on four language pairs (English-German, English-Arabic, English-Russian, and English-Hebrew).\footnote{The qualitative analysis was conducted by one of the authors.} Additionally, we examine word-level results averaged over all languages. Specifically, for each SNC method and language pair, we selected the top and bottom 100 aligned words along with their 10 nearest neighbors in each language. The analysis appears in Section \S\ref{sec:qualitative}.


\section{The Impact of Polysemy on Cross-Lingual Alignment}
\label{app:polysemy}

Polysemy is a linguistic phenomenon where a word has multiple related meanings. For example, the word ``bank'' can refer to a financial institution or the side of a river. 
In linguistics, much research had been conducted on the relation between polysemy and other lexical phenomenon such as rate of change and frequency, including in NLP.
We can differentiate between two types of polysemy: one where a word has multiple related senses, such as ``book'' (a physical object or an act of reserving), and another where the senses are distinct, such as ``bark'' (the sound a dog makes or the outer covering of a tree). 

To manually quantify the polysemy of a word, one approach is to count the number of entries it has in a dictionary. However, this method is not easily applicable to many languages. Alternatively, BabelNet \citep{navigli2010babelnet} can be used to identify word senses. Yet, this introduces challenges, as BabelNet often includes many highly similar senses (that we do not want to take into account), adding noise to our alignment process. While we intend to address these challenges in future work, space constraints and the specific focus of our paper have led us to adopt alternative measures of polysemy for this preliminary inquiry, based on the word representations themselves. Results are presented in Table \ref{app:tab:corr_polysemy_measures}.

\begin{table}[!t]
\centering
\footnotesize
\renewcommand{\tabcolsep}{0.15cm}

\begin{tabular}{cp{0.8cm}p{0.9cm}p{0.9cm}p{0.9cm}p{0.9cm}}
\toprule
& &
{\sc SNC-cloud} &
{\sc SNC-ave} &
{\sc SNC-static} &
{\sc NO} \\ \midrule
\multirow{2}{*}{\textbf{Self-Sim}}
& C & $-0.1^{\star}$ & $-0.3^{\star}$ & $-0.42^{\star}$ & $0$ \\
& D & $-0.41^{\star}$ & $-0.36^{\star}$ & $-0.35^{\star}$ & -$0.37^{\star}$ \\ \midrule
\multirow{2}{*}{\textbf{GMM-Senses}}
& C & $0.16^{\star}$ & $0.22^{\star}$ & $0.31^{\star}$ & $0.21^{\star}$ \\
& D & $0.58^{\star}$ & $0.47^{\star}$ & $0.52^{\star}$ & $0.7^{\star}$ \\ 

\bottomrule
\end{tabular}

\caption{Correlation with Polysemy Measures. Columns represent the two polysemy measures: Self-Similarity (denoted by \textbf{Self-Sim}) and the average number of gmm clusters (denoted by \textbf{GMM-Senses}). word-level correlations are denoted by \textbf{C} and domain-level correlation by \textbf{D}. Neighbors Overlap metric is represented by {\sc NO}. significant correlation with $p < 0.05$ are marked by $\star$. }
\label{app:tab:corr_polysemy_measures}
\end{table}

\subsection{Self-Similarity}

The first measure that we consider is Self-Similarity, introduced by \citep{Ethayarajh19}. This measure have shown to highly correlate with polysemy and was used as an alternative measure for the degree of polysemy of words \citep{gari2021let}. For a word $w$, that is, the average of the pairwise cosine similarities of the representations of its contextualised representations in corpus $\mathcal{A}$. Defined as: 

\begin{equation}
SelfSim = \frac{1}{|\mathcal{A}^2|-|\mathcal{A}|}\Sigma_{i\in \mathcal{A}}\Sigma_{j \in \mathcal{A}, j \neq i}cos(x_{wi},x_{wj}),
\end{equation}  

where $x_{w_i}$ is the contextualised representation of word $w$, taken from its $i$-th instance in corpus $\mathcal{A}$ (here, $|\mathcal{A}| \leq 1000$). For a pair of languages $L_1$ and $L_2$ we calculate the average Self-Similarity score within each language. This average serves as the alignment measure based on Self-Similarity for the language pair.

Table \ref{app:tab:corr_polysemy_measures} shows that at the word-level and the domain-level Self-Similarity is significantly (negatively) weakly-moderately correlated with {\sc SNC} methods, reaching $r=-0.42$ for {\sc SNC-static}. The negative correlations means that the more polysemous the word is, the less it is aligned across languages.

\subsection{Sense Clusters}
We define another measure for polysemy, that is based on clusters of the word embeddings. In the computation process of {\sc SNC-cloud} (Section \S\ref{subsec:alignment_metrics}), for a word $w$ we have $\leq 1000$ contextualised representations, based on an external corpus (Appendix \S\ref{app:sec:exp_setup}), we denote this point-cloud by $\mathcal{O}$. We perform Gaussian-Mixture Models (GMM) clustering on $\mathcal{O}$ and choose the optimal number of clusters using the Elbow Method.\footnote{To ensure robustness, we repeat this process 10 times and select the number of clusters that appears most frequently across these iterations.}

We define the \textbf{degree of polysemy} for a word $w$ to be the number of clusters we calculated. For a pair of languages $L_1$ and $L_2$ we calculate the average degree of polysemy score within each language. This average serves as the alignment measure based on the degree if polysemy for the language pair. 

Table \ref{app:tab:corr_polysemy_measures} shows (the degree of polysemy is denoted by \textbf{GMM-sense}) that the methods are significantly correlated with this measure, at the domain level reaching $r=0.7$ for {\sc NO} and $r=0.58$ for {\sc SNC-cloud}. 

Intuitively, as discussed in the main paper, different patterns of polysemy can result in varied nearest neighbors across languages. For instance, in English, the word \textit{bank} might have a mix of neighbors related to both \textit{river bank} and \textit{financial institution}. However, in Hebrew, where the sense of a river bank does not exist, neighbors are likely all related to the financial institution sense of \textit{bank}. This mismatch can lead to greater divergence in alignment between English and Hebrew.

Although this measure has been shown to reflect the degree of polysemy of words \citep{gari-soler-apidianaki-2021-lets}, the correlations we compute still require further control.


\section{Other Architectures.}
\label{app:sec:other_arch}

\begin{figure}
    \centering
    \includegraphics[width=\columnwidth]
    {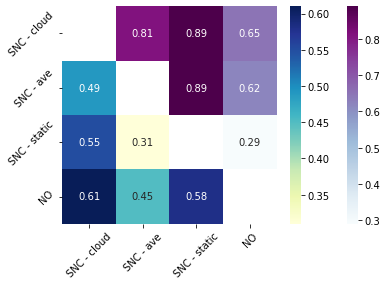}
    \caption{Correlation between the various metrics (other architectures). Pearson correlation is computed for different aggregation methods. The \textbf{upper} matrix represents concept-level correlations, while the \textbf{bottom} matrix represents domain-level correlations. All correlation values are significant with $p<0.05$. }
    \label{app:fig:heatmap_measures}
\end{figure}

We also run our experiments on XLM-RoBERTa-base \footnote{\url{https://huggingface.co/xlm-roberta-base}} for {\sc SNC-cloud} and {\sc SNC-ave} and on 300-dim word2vec multilingual embeddings \footnote{\url{https://github.com/Kyubyong/wordvectors}} for {\sc SNC-static}. Moreover, we run all of the computations for {\sc SNC-cloud} and {\sc SNC-ave} with a different dataset; the Wikipedia section in the Leipzig Corpus, for the latest year available in each language \footnote{\url{https://wortschatz.uni-leipzig.de/en}}. The trends were highly similar to what we report in the main paper and are presented in Table \ref{app:tab:corr_lexical_enviro_other_arch}, Figure \ref{app:fig:box_plot_rob} and Figure \ref{app:fig:heatmap_measures}.

\begin{table}[!t]
\centering
\footnotesize
\renewcommand{\tabcolsep}{0.15cm}
\begin{tabular}{cp{0.8cm}p{0.9cm}p{0.9cm}p{0.9cm}p{0.9cm}}
\toprule
& &
{\sc SNC-cloud} &
{\sc SNC-ave} &
{\sc SNC-static} &
{\sc NO} \\ \midrule
\multirow{2}{*}{\textbf{CLT}}
& C & $0.1^{\star}$ & $0.08$ & $0.27^{\star}$ & $0$ \\
& D & $0.23^{\star}$ & $0.31^{\star}$ & $0.11^{\star}$ & $0.08^{\star}$ \\ \midrule
\multirow{2}{*}{\textbf{GEO}}
& C & $0.1$ & $0.08^{\star}$ & $0.15^{\star}$ & -$0.01$ \\
& D & $0.2^{\star}$ & $0.39^{\star}$ & $0.1^{\star}$ & $0.17^{\star}$ \\ \midrule
\multirow{2}{*}{\textbf{frequency}}
& C & $0$ & $-0.04$ & $0.01$ & $0$ \\
& D & $0.35^{\star}$ & $0.15^{\star}$ & $0$ & $0.58^{\star}$ \\ \midrule
\multirow{2}{*}{\textbf{concreteness}}
& C & $0$ & $0$ & $0$ & $0.1$ \\
& D & $0.15^{\star}$ & $0.1^{\star}$ & $0.15^{\star}$ & $0.08$ \\ \midrule
\multirow{2}{*}{\textbf{rate-change}}
& C & -$0.25^{\star}$ & -$0.27^{\star}$ & -$0.3^{\star}$ & -$0.11$ \\
& D & -$0.55^{\star}$ & -$0.48^{\star}$ & -$0.65^{\star}$ & -$0.39^{\star}$ \\
\bottomrule
\end{tabular}

\caption{Correlation with lexical and enviromental features (other architectures). Columns represent the features (CLT denotes cultural distance and GEO denotes geographical distance) and subcolumns represents concept-level aggregation (C) vs. domain-level aggregation (D). {\sc NO} represents Neighbors Overlap metric. significant correlation with $p < 0.05$ are marked by $\star$. }
\label{app:tab:corr_lexical_enviro_other_arch}
\end{table}


\end{document}